# ENHANCEMENT PERFORMANCE OF ROAD RECOGNITION SYSTEM OF AUTONOMOUS ROBOTS IN SHADOW SCENARIO

Olusanya Y. Agunbiade[1], Tranos Zuva[1], Awosejo O. Johnson[2], Keneilwe Zuva[3]

[1]Department of Computer System Engineering, Tshwane University of Technology, Pretoria, South Africa
[2]Department of Business Informatics, Tshwane University of Technology, Pretoria, South Africa
[3]Department of Computer Science, University of Botswana, Gaborone, Botswana

## ABSTRACT

*Road region recognition is a main feature that is gaining increasing attention from intellectuals because it helps autonomous vehicle to achieve a successful navigation without accident. However, different techniques based on camera sensor have been used by various researchers and outstanding results have been achieved. Despite their success, environmental noise like shadow leads to inaccurate recognition of road region which eventually leads to accident for autonomous vehicle. In this research, we conducted an investigation on shadow and its effects, optimized the road region recognition system of autonomous vehicle by introducing an algorithm capable of detecting and eliminating the effects of shadow. The experimental performance of our system was tested and compared using the following schemes: Total Positive Rate (TPR), False Negative Rate (FNR), Total Negative Rate (TNR), Error Rate (ERR) and False Positive Rate (FPR). The performance result of the system improved on road recognition in shadow scenario and this advancement has added tremendously to successful navigation approaches for autonomous vehicle.*

## KEYWORDS

## 1. INTRODUCTION

Road region is an important feature that is necessary for an autonomous driving system to detect for navigation. Various techniques have been used by previous researchers to detect road region but recently vision based system is preferred to be used because of the ability to acquire data in a non-intrusive way[1].

Based on these properties, some researchers used vision based technique to extract lane feature for road region recognition because lane has characteristics that are used for collision prevention, lane departure warning, vehicle navigation and lateral control [2]. Others used the vision based technique for the extraction of colour, because colour has a distinctive characteristic that can be used for classification. The assumption that road region has similar colour different from non-road region can be used for the recognition of road area [3]. While some focus on using color technique for the extraction of edges. Edges extraction has been used for boundary detection which is used for classification of image into road region and non-road region [4].

Furthermore, for a more robust system based on vision based technique, some researchers combined two extracted features together for the recognition process of road region. Colour and texture was used in [5, 6] for the detection of road region. Irrespective of all these techniques, shadow limits the performance of the system because of the capability of corrupting the feature extraction and leads to misclassification of road region as non-road region and vice versa. In this paper, we introduce into the road region recognition system an algorithm capable of addressing the effect of shadow. As a result, this algorithm improves the road region recognition performance of the autonomous vehicle.

The rest of is paper is structured as follows: Section II review the related work of past researcher, in section III, analysis review of the methodology used for the road region recognition of the system. Section IV demonstrates flows of pool of pseudo code used for the road recognition system. Results and assessment scheme were revealed in section V and the research conclusion was done in section VI.

## 2. LITERATURE REVIEW

Road region is a significant issue to consider for a successful movement of an autonomous vehicle across a long range distance. Vision-based method has been a common technique used by various researchers for developing an autonomous vehicle and has been productive towards road region recognition.

In [7, 8], the intelligent vehicle region detection system uses colour extraction for road region recognition. At first, one dimensional colour histogram feature is set up and thresholding is applied for grouping similar pixels for the classification process. However, when shadow is present, occupied shadow area of road region and non-road region have very comparable features and at this point misclassification of road region as non-road region and vice versa can occur.

In [9], colour and edge are used for the detection of road region on a low resolution image. They used a low resolution image because it makes road surface colour seem to be consistent making it easier when using colour for extraction of road region. However, boundary wearing makes colour not suitable for boundary extraction so Bezier spline edge detection algorithm with control point optimization as expressed in equation (1) was introduced to the system to extract the edge that best fit the road boundary. Based on this approach, remarkable result better than using colour extraction is achieved but complain illuminative effect (shadow and light intensity) limiting the system and will be addressed in their future work [9]

$$P(t) = (1-t)^3 P_0 + 3t(1-t)^2 P_1 + 3t^2(1-t)P_2 + 6t^3 P_3 \tag{1}$$

$$= (t^3, t^2, t^1, 1) \begin{pmatrix} -1 & 3 & -3 & 1 \\ 3 & -6 & 3 & 0 \\ -3 & 3 & 0 & 0 \\ 1 & 0 & 0 & 0 \end{pmatrix} \begin{pmatrix} P_0 \\ P_1 \\ P_2 \\ P_3 \end{pmatrix}$$

Where $P_0, P_1, P_2, P_3$ are the control point which requires the Bezier spline to pass through $t \in [0,1]$

In [10], colour information is used for extraction, but first RGB colour properties of the image are converted into shadow-invariant feature space and also model-based classifier was introduced to the system for optimising the classification process on the other hand since there road detection approach is colour based, there system can still fail under severe lighting variation [6]. In this paper, since our approach is also coloured based, we optimized by introducing to the road region recognition system an algorithm based on normalized differences index (NDI) and morphological operation to enhance the system against the effect of shadow.

## 3. METHODOLOGY

The vision method for road region recognition is based on four stages, The feature extraction which is based on colour is used for image analysis, filtering stage which was introduced to enhance the system by addressing the effect of shadow for proper classification to happen at the segmentation stage. Post processing was used to improve the classification result at the segmentation operation for proper navigation of autonomous vehicle

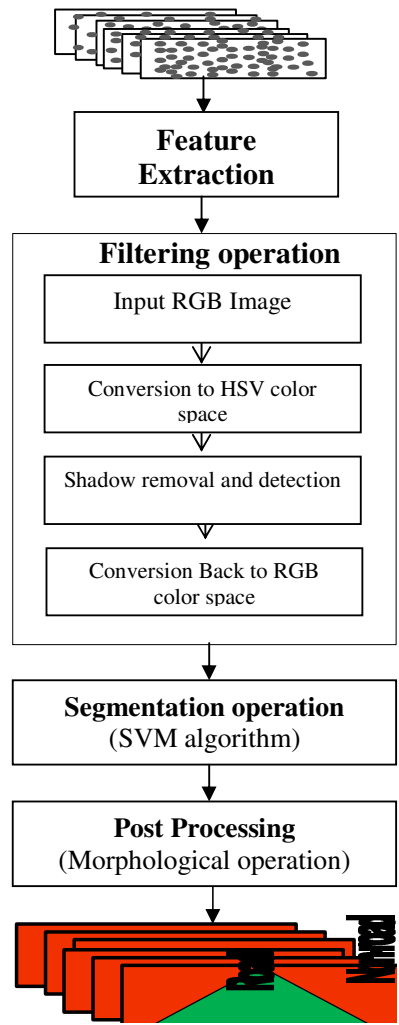

Figure 1 Stages of Road Recognition

### 3.1. Feature Extraction Operation

Feature extraction is a vital stage in image processing and at this stage various techniques such as texture and edge can be used for road region recognition. However, colour is also a common technique used by previous researchers [3, 8, 11], as it holds important information for objection identification in all images. Thus, the assumption that road region would be more-or-less consistent in their mixture and road region in RGB space is brighter than others are colour based extraction used in [3] for road region classification. In this paper, basis of colour similarity was used for road region recognition and Mahalanobis distance $d$ expressed in equation (2) was employed for similarity extraction because of better approximation than Euclidean distance [12].

$$d = \sqrt{(m_t - x)^t \sum\nolimits_t^{-1} m_t - \sum\nolimits_t^{-1} x} \qquad (2)$$

Where x denote studied pixel, $m_t$ represent the mean matrix of the training set and $\sum\nolimits_t$ signify the covariance matrix of the training set.

### 3.2. Filtering Operation

Filtering Operation is used to optimize road region recognition system. It plays a significant operation to sharpen and supress image noise since the environmental noise (shadow) has already altered the RGB colour value used for road region recognition. Various road recognition system designed by different researchers failed when encountering shadow and some of these systems are studied in our literature review. However, some researchers tried to solve this problem using invariant HSV colour space but failed under severe shadow effect [10].

In this research, Filtering algorithms introduced at this stage are capable of detecting and eliminating the effect shadow to produce a corrupt free RGB colour value of the image. This algorithm is based on Normalised Index Differences (NDI) and morphological operation. The algorithm proposed, first converts RGB colour value of the image to HSV using equation (3)-(5), because in HSV colour space, shadow holds spectral properties for easy identification [13].

$$V = \frac{(R + G + B)}{3} \qquad (3)$$

$$S = 1 - \frac{3}{R + G + B} \min(R, G, B) \qquad (4)$$

$$H = \begin{cases} \theta & \text{if } B \leq G \\ 360^0 & \text{if } B \leq G \end{cases} \qquad (5)$$

$$\text{where } \theta = \cos^{-1}\left\{ \frac{\frac{1}{2}[(R - G) + (R - B)]}{\sqrt{(R - G)^2 + (R - B)(G - B)}} \right\},$$

R, G, B represent red, blue, green respectively

The algorithm process for is based on using the saturation and value component of the image are used for shadow region extraction as expressed in equation (6)

$$NDI = S - V/S - V \tag{6}$$

where S and V represent respectively saturation and value

OTSU threshloding algorithm is used to find an optimal threshold (T) that will be used to label pixels and for partitioning the NDI image as illustrated in equation (7) for easier analysis.

$$NDI(T) = \left(\bar{\mu}.w(T) - \mu(T)\right)^2 / w(T).\mu(T) \tag{7}$$

where $w(T) = \sum_{i=0}^{T} p_j$, $\mu(T) = \sum_{j=T+1}^{255} p_j$, $\bar{\mu} = \sum_{i=0}^{255} j.p_j$,

$p_j$ represent the probability of pixel with grey level j in the image.

The pixels of an image having higher NDI than the threshold (T) are labelled shadow pixel otherwise non-shadow as shown in equation (8).

$$I_{shadow}(i, j) = \begin{cases} 1 & NDI(i, j) \geq T \\ 0 & NDI(i, j) \langle T \end{cases} \tag{8}$$

$I_{shadow}(i, j)$ represent the binary image obtained after thresholding (T) with the set of shadow pixel set to 1 and non-shadow pixel set to 0.

In the process of shadow elimination, connecting algorithm used for shadow classification is a morphological operation used to connect region label has 1. Equation (9) shows the illustration of the connected component.

$$I_m = (I_{m-1} \oplus B) \cap I_{shadow} \qquad m = 1, 2, 3, \ldots \tag{9}$$

where B represents the structuring element that terminate when $I_m = I_{m-1}$. $I_m$ denote connected component of $I_{shadow}$

The expression in equation (9) generates many sets of 'n' connected component of different shadow that exists in an image. In the elimination process, the next operation is the buffer area estimation which is the non-shadow area around that shadow area [13]. Equation (10) shows the illustration of buffer area estimated using the morphological dilation process and image

$$I_{buff,k} = (I_{dilated,k} - I_k) \tag{10}$$

thus,

$$I_{dilated,k} = (I_k \oplus \mathbf{B}_{square})$$

where $I_{buff,k}$ shows the location of the non-shadow points around the shadow area, $I_{dilated,k}$ represent the morphological dilation operation that will expand the shadow boundaries, $I_k$ denote the each shadow area of the image, $\mathbf{B}_{square}$ denote 3x3 square structure element used for the dilation process and k = 1, 2, 3, 4….n

In the eliminating process of shadow area, the mean and the variance of the non-shadow area (buffer area) are used to compensate the shadow area. Equation (11) shows the illustration

$$I_k^{'}(i,j) = \mu_{buff,k} + \left(\frac{I_k(i,j) - \mu_k}{\sigma_{buff,k}}\right) \cdot \sigma_k \tag{11}$$

where $\mu_{buff,k}$ and $\sigma_{buff,k}$ respectively signify mean and variance of the pixels of image I at location $I_{buff,k}$; $\mu_k$ and $\sigma_k$ are the mean and variance of the shadow pixel of image I at location $I_K$

### 3.3. Segmentation Stage

This stage is essential because the segmentation algorithm performs a vital task, since the road classification at the feature extraction stage is in-proper due to corrupted RGB colour value caused by shadow [11]. However, better classification and detection of road recognition is achieved at the segmentation stage than feature extraction stage, because the RGB colour value at this stage has been processed by the filtering algorithm.

Various segmentation algorithms exist, but Support Vector Machine (SVM) has been a common technique used by various researchers [14, 15] and it can be used for both regression and pattern recognition [15]. The idea for classification used by SVM is to find the hyperplane expressed in equation (12) that will be used for class separation

$$H = w..x + b \tag{12}$$

where $x$ represents points on the hyper plane, $w$ signifies a $n$-dimensional vector perpendicular to the hyper plane, b denote the distance of the closet point on the hyper plane.

The hyperplane is defined as the distance that exists between neighbouring vectors for two classes [14].

Equation (13) shows the formulation of the problem.

$$\text{Minimize } \frac{1}{2}\|w\|^2, \text{ subjected to constraint } y_i(wx_i + b) - 1 \geq 0 \quad \forall i \tag{13}$$

where $\frac{1}{2}\|w\|^2$ is the objective function, $i$ denote image pixel.

Applying the Lagrange multiplier, the optimization problem can be converted into quadratic programming problem expressed in equation (14)

$$\underset{\lambda_1 \ldots \ldots \lambda_l}{\text{Maximize}} \quad \sum_{i=1}^{l} \lambda_i - \frac{1}{2}\sum_{i=1}^{l}\sum_{i=1}^{l} \lambda_i \lambda_j y_i y_j x_i .. x_j \tag{14}$$

subjected to constraint

$$\sum_{i=1}^{l} \lambda_i y_i = 0, \quad \lambda_i \geq 0 \quad i = 1 \ldots \ldots \ldots l$$

Signal & Image Processing : An International Journal (SIPIJ) Vol.4, No.6, December 2013

The SVM solution is illustrated in equation (15)

$$w = \sum_{i=1}^{l} \lambda_i y_i x_i \quad , \quad b = y_i - w.x_j \tag{15}$$

The decision function used for classification and pattern recognition in SVMs is expressed in equation (16)

$$f(x) = \text{sign}(w.x + b) = \text{sign}\left(\sum_{i=1}^{l} y_i \lambda_i (x . x_i) + b\right) \tag{16}$$

where $\lambda_1........\lambda_l$ is the vector of non-negative Lagrange multiplier corresponding to constraint in (13); the vector $x_i$, $y_i$ for which $\lambda_i \rangle 0$ are called support vector and other training vectors have $\lambda_i = 0$ [14].

### 3.4. Post Processing

Post processing is an important and common technique used in image processing. Several post processing exist. However in our post processing, morphological operation was employed because of impressive result in image analysis [16]. Morphological operation is based on the assumption that images consists of structures which may be handled by set theory [17]. However, based on the execution of set theory assumption, images largest and connected areas are extracted and labelled as road part and other parts different from road part are labelled as non-road. In the drivable part recognition system, the morphological operation was introduced to improve and achieve an enhanced classification result for more perfect autonomous vehicle navigation than previous stages. In morphology, operation based on the following are involved:

Image dilation as illustrated in equation (17) is executed by laying structure element B on image set M. These allow Road expansion because it fills existing small holes and connect disjoint object.

$$M \oplus B = \bigcup_{b \in B} M_b \tag{17}$$

Image erosion as illustrated in equation (18) is a related operation to dilation but it tries to shrink by eroding away their boundary.

$$M \ominus B = \{p | p + b \in M \; \forall b \in B\} \tag{18}$$

Image dilation and erosion are the main operations in morphology, but filters illustrated in equation (19) and (20) are also important operations to consider as they helps to smoothing previous operations of dilating and eroding [17].

$$\text{Morphological opening: } M \circ B = (M \ominus B) \oplus B \tag{20}$$
$$\text{Morphology closing: } M \bullet B = (M \oplus B) \ominus B \tag{21}$$

The road edge of set image M represented by $\beta(M)$ is extracted by performing the set differences between $M$ and erosion of $M$ by $B$ as expressed in (22)

$$\beta(M) = M - (M \ominus B) \tag{22}$$

We proposed a simple algorithm for road recognition using region filling based on image dilation, complement and intersection as illustrated in equation (23). Starting with point J inside the boundary. The procedure then fills respectively the color allocated to road region for better classification [18].

$$N_k = (N_{k-1} \oplus B) \cap M^c \qquad k = 1, 2, 3, \ldots\ldots \tag{23}$$

Where $N_0 = J$ (starting point), B is symmetric structure element, $M^c$ is the complement of set image M, at iteration step k when $N_k = N_{k-1}$ the algorithm terminates [18].

## 4. POOLS OF PSEUDO CODE

### 4.1. Pseudo Code for Road Region Recognition Feature Extraction

Input: select frame M from stream 1……P images
Output: extracted RGB value of M from stream 1….P images
Step 1: selected M from 1…….P images
Step 2: extracting color value M based on RGB color space
Step 3: apply (2) on M from stream 1…..P images for similarity extraction
Step 4: extract road region of M from 1…..P images

In Phase one, Mahalanobis distance proposed for road pattern recognition is based on the assumption that road area has similar color different from non-road [3]. However system based on this assumption failed when facing shadow.

### 4.2. Pseudo Code for Shadow Operation

Input: RGB extraction of stream 1……P images
Output: uncorrupted RGB extraction of stream 1…..P images

Step 1: select M (corrupted or uncorrupted depending on shadow presence) from 1….P images.
Step 2: apply equation (3) - (5) on M from stream 1…..P images.
Step 3: apply equation (6) – (10) on M from stream 1……P images for shadow detection.
Step 4: apply equation (11) on M from stream 1…….P images for shadow removal.
Step 5: conversion of HSV color space of M from stream 1…..P images to RGB.
Step 6: extracted uncorrupted RGB value of M from 1…..P images.

In phase 2, since the first phase of system is affected by shadow, filtering algorithm proposed here is just to address the effect of shadow for proper road recognition to be achieved at the next phase.

### 4.3. Pseudo Code for Segmentation Algorithm.

Input: stream 1…..P images with uncorrupted RGB value
Output: road pattern recognition of stream 1…… P images

Step 1: select M (uncorrupted RGB value) from stream 1…..P images
Step 2: apply equation (13) on M from stream 1…….P images for linear separable.
Step 3: apply equation (14) on M from stream 1……..P images to obtain quadratic programming problem.
Step 4: road pattern recognition on M from stream 1…..P images: apply equation (16)
Step 5: extracted road and non-road of M from 1…P images.

Signal & Image Processing : An International Journal (SIPIJ) Vol.4, No.6, December 2013

In phase 3, SVM achieved better result than feature extraction because the shadow effect at this stage has been eliminated. SVM classification is done by trying to find a hyperplane and it was proposed because of good generalization capability.

### 4.4. Pseudo Code for Morphological Operation

Input: road pattern recognition
Output: enhanced road pattern recognition of stream 1…...P images

Step 1: select M (road pattern based on SVM) from i……. P images
Step 2: binary dilation of M from 1…..P images apply equation (17)
Step 3: binary erosion of M from 1……P images apply equation (18)
Step 4: apply morphological filters on M from stream 1…..p images to smoothing previous operation of equation (17) and (18)
Step 5: boundary detection of M from stream 1….P images apply equation (22)
Step 6: enhanced road recognition of M from 1…..P images is extracted using equation (23).

In phase 4, morphological operation based on the assumption that road area are connected [15] was used to accomplish better classification result than support vector machine and this enhancement has also improved the navigation of autonomous vehicle.

## 5. RESULT AND ASSESSMENTS

In this research, the optimized road recognition system performance evaluation was examined first qualitatively by visual comparison between the real world image captured by the camera and vision output image of the autonomous robot

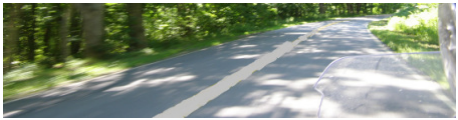 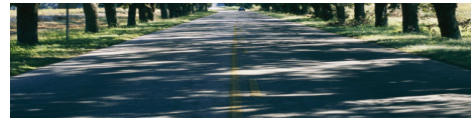

Figure 2a input image captured by camera

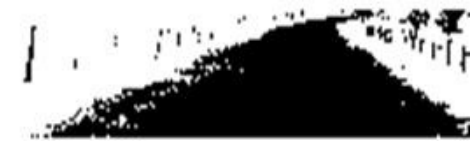 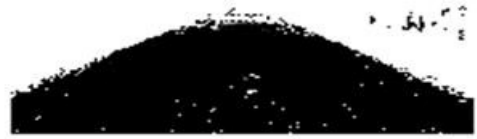

Figure 2b output image of the autonomous vehicle

The evaluation schemes used to test the output image of the system with filtering algorithm are expressed in equation (24)–(29). These evaluation schemes illustrate the measureable rate of how successful the optimized road recognition system has improved the navigation of the autonomous vehicles.

The ACC signified as accuracy rate (24), the section of total numbers of roads and non-road that are predicted properly. ERR symbolize error rate (25), the section of total numbers of roads and non-road that are predicted erroneously. TPR is the total positive rate (26), the section of roads case that are properly classified. FNR denoted as the false negative rate (27), the section of roads case that are erroneously categorized as non-road. TNR represent total negative rate (28), the section of non-roads case that are categorized correctly. FPR is characterized as false positive rate (29), the section of non-roads cases that are categorized as road.

$$\text{ACC} = \frac{TP_i + TN_i}{N_i} \tag{24}$$

$$\text{ERR} = \frac{FN_i + FP_i}{N_i} \tag{25}$$

$$\text{TRP} = \frac{TP_i}{TP_i + FN_i} \tag{26}$$

$$\text{FNR} = \frac{FN_i}{TP_i + FN_i} \tag{27}$$

$$\text{TNR} = \frac{TN_i}{TN_i + FP_i} \tag{28}$$

$$\text{FPR} = \frac{FP_i}{TN_i + FP_i} \tag{29}$$

where $TP_i$ symbolize true positive: number of road pixel properly categorized in the $i^{th}$ of A; $TN_i$ is represented as true negative: non-road pixel that are properly classified in the $i^{th}$ of A; $FP_i$ denote false positive: numbers of non-road pixel categorized as road in the $i^{th}$ of A; $FN_i$ Signifies false negative: numbers of road pixel categorized as non-road in the $i^{th}$ of A; $N_i$ indicate number of non-road pixel and road pixel in $i^{th}$ ground truth classification.

Based on 100 frames of road image, the assessment illustrated above was used to test on an average based on 10 frames for error rate and accuracy detection. Comparison result of the system with filtering algorithm and vice versa was done. This result proves better performance of the system proposed. Illustration of the comparison result is expressed in figure (3).

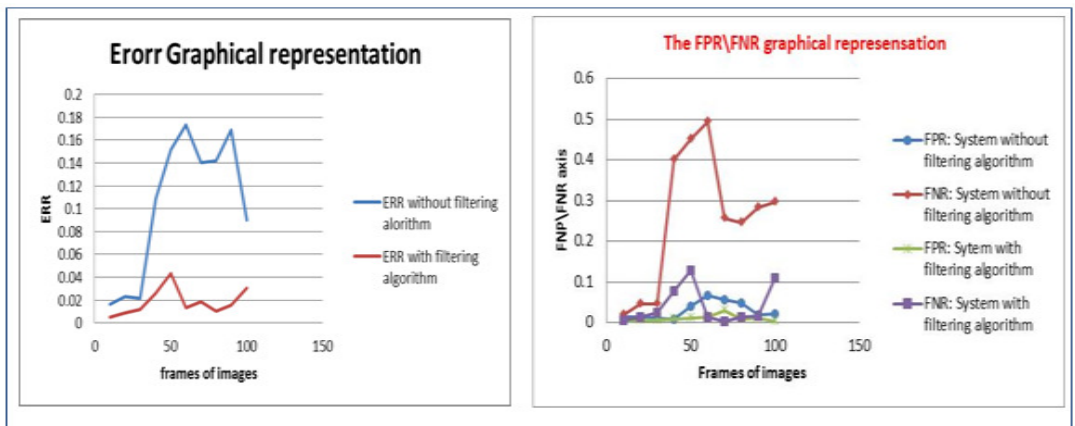

Fig 3a shows error rate illustration     Fig 3b shows FP/FN error rate illustration

Figure 3: Result of the Experiment

The comparison representation based on ERR, FRP, and FNR was done to check the system performance when filtering algorithm is not introduced to the system and vice versa.

## 6. CONCLUSION

In this research, a road recognition method was proposed for out-door autonomous vehicle for proper navigation in the presence of shadow. The shadow algorithm based on NDI and morphological operation was used for proper segmentation and classification of image pixel. This helped in the application process of shadow removal algorithm because distinctive area affected by shadow effect is well classified. The algorithm is efficient because it has improved the performance of road recognition system by removing the effect of shadow for proper classification to be achieved at the segmentation stage. Improving the segmentation stage is the involvement of morphological operation to achieve better classification result for road recognition. The system was tested and proven with various roads with shadow and results show autonomous robot achieving proper navigation.

## ACKNOWLEDGEMENT

The authors would like to appreciate the effort made by Tshwane University of Technology towards this work by providing funds and required resources. It would have been impossible to complete this research without their support.

**AUTHOR**

Agunbiade Olusanya Yinka is a B-TECH graduate of Ladoke Akintola University of Technology in Computer Science and presently, a master student in the Department of Computer System Engineering at Tshwane University of technology. His focus area lies in Image Processing, Computer Vision, Data Mining, Ubiquitous Intelligence, Artificial Intelligence Numerical Computation and Knowledge Management.

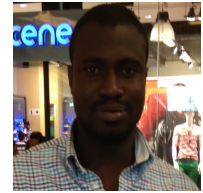